\pdfoutput=1

\documentclass[11pt]{article}

\usepackage{EMNLP2023}

\usepackage{times}
\usepackage{latexsym}
\usepackage{subcaption} 
\usepackage{graphicx}
\usepackage[shortlabels]{enumitem}

\usepackage[T1]{fontenc}

\usepackage[utf8]{inputenc}

\usepackage{microtype}

\usepackage{inconsolata}

\title{The Law and NLP: Bridging Disciplinary Disconnects}

\author{Robert Mahari \\
  MIT and Harvard Law School \\
  \texttt{rmahari@mit.edu} \\\And
  Dominik Stammbach \\
  ETH Zurich \\
  \texttt{dominsta@ethz.ch} \\ \AND
  Elliott Ash \\
  ETH Zurich \\
  \texttt{ashe@ethz.ch} \\\And
  Alex `Sandy' Pentland \\
    MIT \\
\texttt{pentland@mit.edu} \\
  }

\begin{document}
\maketitle
\begin{abstract}
Legal practice is intrinsically rooted in the fabric of language, yet legal practitioners and scholars have been slow to adopt tools from natural language processing (NLP). 
At the same time, the legal system is experiencing an access to justice crisis, which could be partially alleviated with NLP.
In this position paper, we argue that the slow uptake of NLP in legal practice is exacerbated by a disconnect between the needs of the legal community and the focus of NLP researchers.
In a review of recent trends in the legal NLP literature, we find limited overlap between the legal NLP community and legal academia.
Our interpretation is that some of the most popular legal NLP tasks fail to address the needs of legal practitioners. 
We discuss examples of legal NLP tasks that promise to bridge disciplinary disconnects and highlight interesting areas for legal NLP research that remain underexplored.
\end{abstract}

\section{Introduction}

Rapid advances in NLP technology are already promising to transform society and the economy~\citep[see e.g.,][]{openai2023gpt4, eloundou2023gpts, bommasani2021opportunities}, not least by their impact on many professions. Given that legal practice is embedded in written language, it stands to gain immensely from the application of NLP techniques. 
This promise has attracted research on NLP applications related to a wide array of legal tasks including legal research~\citep{huang2021context, ostendorff2021evaluating}, legal reasoning~\citep{guha2023legalbench, mahari2021autolaw}, contract review~\citep{hendrycks2021cuad, leivaditi2020benchmark}, statutory interpretation~\citep{nyarko2022statistical,savelka2019improving}, document review~\citep{yang2022goldilocks, zou2020towards}, and legal question answering~\citep{vold2021using, khazaeli-etal-2021-free, martinez2023survey}.

The current model of legal services is failing to address legal needs in several important contexts.
In the United States, around 92\% of civil legal problems experienced by low-income Americans receive no or inadequate legal help~\citep{LegalServicesCorporation2022}.
In U.S. criminal cases, where criminal defendants generally have a right to legal counsel, public defenders are systematically overworked and under-resourced~\citep{Pace2023}.
Access to legal services is also limited for U.S. small businesses~\citep{baxter2022dereliction}. 
The combination of unequal access to and high costs of legal services results in a troubling access to justice issue.

Internationally, there is tremendous variation in legal systems and challenges, but many opportunities where technology could help address legal inefficiencies and under-served communities have been identified globally~\citep[see e.g.,][]{wilkins2017indian, cunha2018brazilian, bosio2023survey, WorldJusticeProject2019}. 
Against this backdrop, numerous scholars have focused on legal technology as a path towards reducing the access to justice gap by extending the abilities of legal professionals and helping non-lawyers navigate legal processes~\citep{baxter2022dereliction,bommasani2021opportunities, cabral2012using, rhode2013access, katz2021legal}.  

Of course, technology is no panacea and many of the inequities and inefficiencies in jurisprudence are related to deeply rooted social and cultural phenomena.
Nonetheless, we view the \emph{responsible} deployment of legal technology as a crucial step towards improving the legal profession and broadening access to legal services.

Despite their potential to improve the provision of legal services, the legal industry has been slow to adopt computational techniques. 
The majority of legal services continue to rely heavily on manual work performed by highly trained human lawyers.
This slow adoption may be partially attributed to risk aversion, misaligned incentives, and a lack of expertise within the legal community~\citep{Livermore_Rockmore_2019,fagan2020natural}.

We argue that there is another factor to blame, rooted not in legal practice but rather in legal NLP research:
In short, legal NLP is failing to develop many applications that would be useful for lawyers. 
Instead, legal NLP research tends to focus on generic NLP tasks and applies widely-used NLP methodologies to legal data, rather than developing new NLP tools and approaches that solve problems unique to the legal context.

For example, NLP research might apply text classification to predict the direction of a U.S. Supreme Court judgment based on portions of the judicial opinion.
These types of models tend to be of limited practical utility: First, the vast majority of lawyers and legal disputes will never reach the Supreme Court. Second, the legal reasoning applied by the Supreme Court is unlikely to be representative of lower courts. And lastly, classifiers trained on published judgments may emulate judicial idiosyncrasies rather than modeling optimal legal reasoning. 
Meanwhile, there has been less research on systems to help lawyers identify relevant precedent for trial, on exploring automated summarization and generation of legal documents, or leveraging NLP for online dispute resolution.

The work done by \citet{Livermore_Rockmore_2019,katz2021legal, cabral2012using, rhode2013access} and others takes an important step toward bridging disciplinary disconnects by providing overviews of NLP and related methods to legal and multidisciplinary communities.
We hope to build on this work by encouraging the legal NLP community to understand the needs of legal practitioners. 
Our paper offers some initial starting points for NLP research that are informed by practical needs.

\begin{figure}
    \centering
    \includegraphics[width=\linewidth]{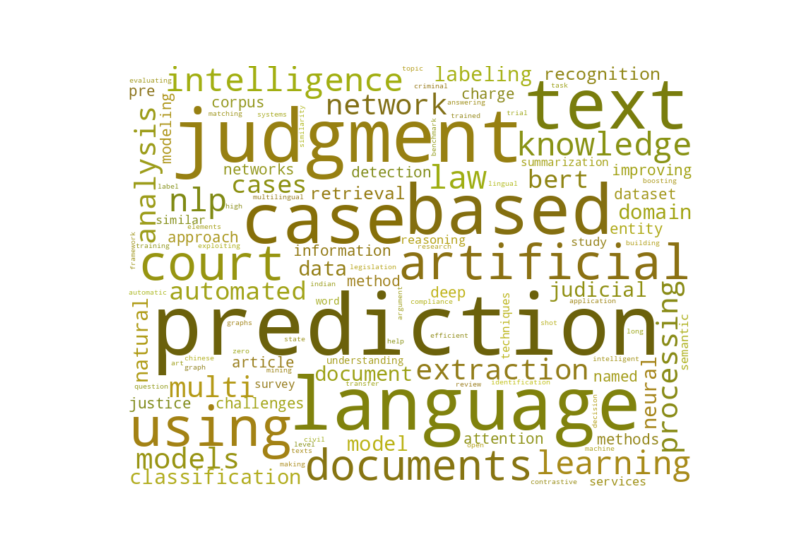}
    \caption{Word cloud of legal NLP paper titles.}
    \label{fig:word_cloud}
\end{figure}

We base our argument on a review of recent legal NLP research, which identifies key themes in this literature (see Figure \ref{fig:word_cloud}). 
We find that a large portion of this research focuses on tasks which we believe are disconnected from the needs of legal practitioners. 
We further observe that only a small fraction of citations to legal NLP publications stem from legal publications, providing evidence that NLP publications have not managed to rise to the attention of the legal community (see left panel of Figure \ref{fig:main_results}). 
Grounded in this review, we segment legal NLP tasks into three categories: applications that could aid the provision of legal services; widespread NLP applications that have limited impact on practical legal issues; and areas of legal NLP research that could have significant impact on legal practice but which remain underexplored. 

Our work adds to a growing number of recent position papers discussing the intersection of law and NLP~\citep[see e.g.,][]{dale_2019, zhong-etal-2020-nlp, tsarapatsanis-aletras-2021-ethical, deOliveira2022, katz2023natural}. 
The number of survey papers in this domain might suggest some confusion about the state of legal NLP. 
In addition to offering descriptive findings, we aim to provide a normative argument. 
In light of the access to justice issues highlighted above, we encourage legal NLP researchers to pragmatically focus on work that promises to broaden access to justice.
This objective helps advance a shared normative goal that does not neglect the `legal' dimension of legal NLP.

To summarize, we make the following contributions and recommendations.

\begin{enumerate}[(1),left=0mm,topsep=0.1mm,noitemsep]
\item We review the current state of legal NLP.
\item We discuss underexplored areas of legal NLP research.
\item We propose the use of legal NLP to tackle the access to justice crisis as a shared normative goal.
\item We advocate for more collaboration between NLP researchers and legal scholars and practitioners.
\end{enumerate}

\section{Literature Review}

\begin{figure*}[htpb]
\centering
\hspace{-10mm}
\begin{subfigure}[t]{0.45\textwidth}
    \includegraphics[width=\linewidth]{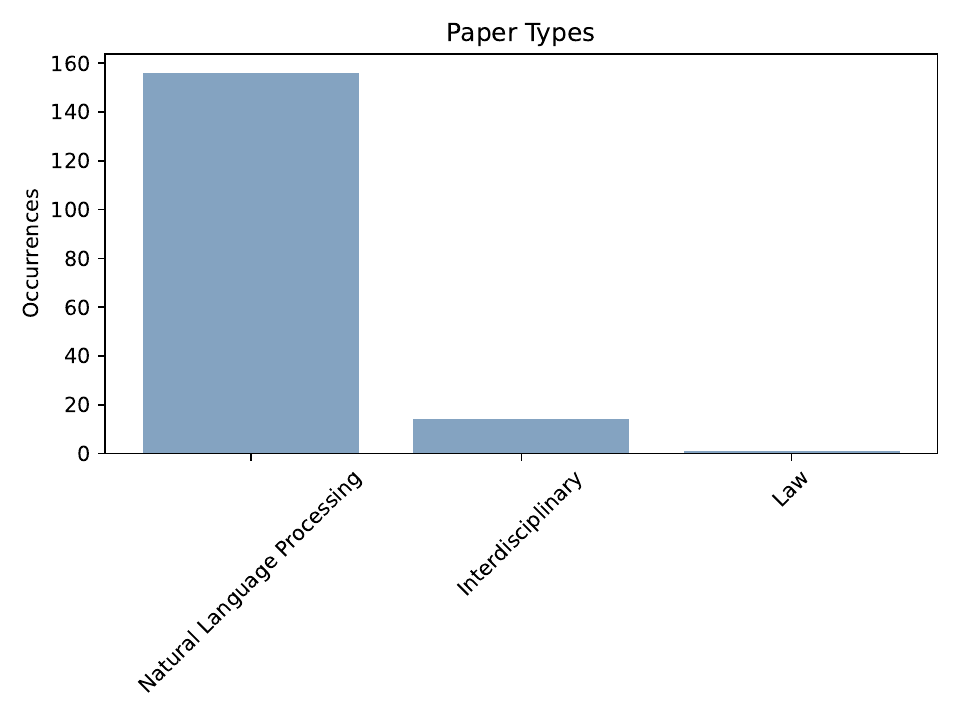}
    \label{fig:paper_types}
\end{subfigure} \hspace{-2mm}
\begin{subfigure}[t]{0.45\textwidth}
    \includegraphics[width=\linewidth]{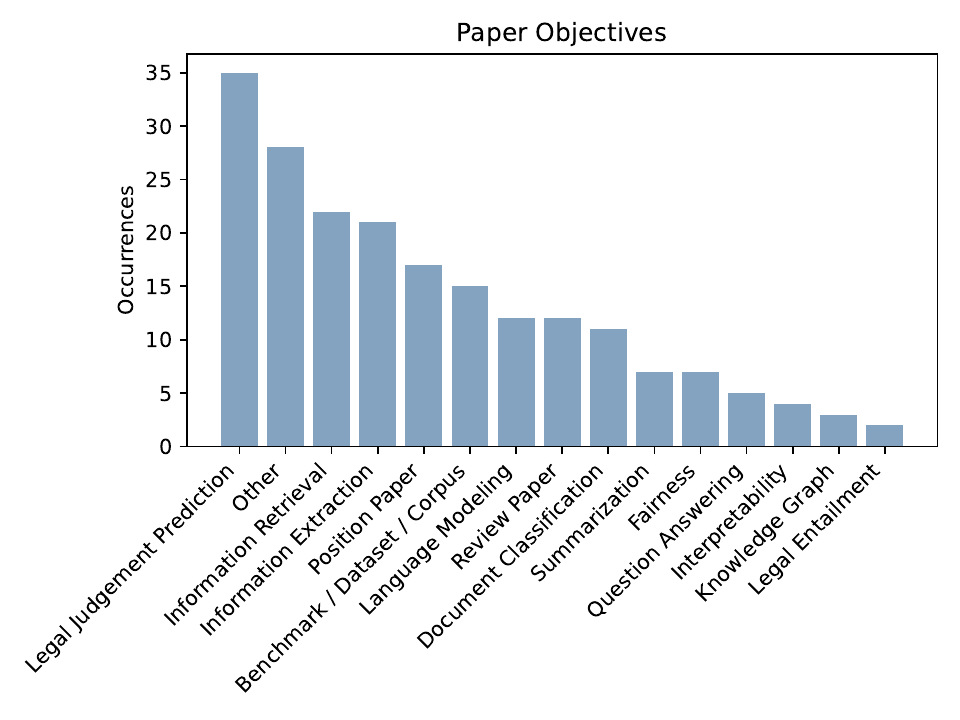}
    \label{fig:objectives}
\end{subfigure} \hspace{-2mm}
\hspace{-10mm}
\vspace{-6mm}
\caption{(Left) Bar plot of paper categories in our literature review (y-axis is the total number of papers of each type). (Right) Bar plot of objectives in the reviewed papers (y-axis is the total number of papers for each objective).}
\label{fig:main_results}
\vspace{-5mm}
\end{figure*}

We conduct a rapid literature review via forward citation chasing over 171 papers, following recommendations made by~\citet{Khangura2012}. Literature reviews have been employed in several position papers investigating the intersection of NLP and other disciplines~\citep[see e.g.,][]{doogan_review, slr_application}.

The starting point of our rapid review is \citet{zhong-etal-2020-nlp}, an overview paper about how legal NLP benefits the legal system. 
We selected this work as our starting point because it is a \textit{recent} contribution in a top NLP conference that has received a reasonably high number of citations for us to review. 
Moreover, it provides (to the best of our knowledge) a fairly accurate overview about the current state of legal NLP. 
Our review includes all papers citing~\citet{zhong-etal-2020-nlp} on Google Scholar. 

\citet{zhong-etal-2020-nlp} aim to identify popular legal NLP tasks and discuss these tasks from the perspectives of legal professionals. 
Tasks outlined in the paper include legal judgment prediction, question answering, and similar case matching. 

We read and manually annotate all papers citing~\citet{zhong-etal-2020-nlp}, resulting in 171 annotated papers.\footnote{At the time of writing, the paper had 182 citations in total. We discarded nine citations for which Google Scholar did not provide a link.} 
For each paper, we determine:

\begin{enumerate}
[(1),left=0mm,topsep=0.1mm,noitemsep]
\item Is the paper is mainly a NLP contribution, an interdisciplinary contribution, or a legal contribution?
\item What are the paper's key objectives (e.g., legal judgment prediction, summarization, position paper, introducing a benchmark)? A paper can have multiple objectives.
\end{enumerate}

We follow a hybrid inductive-deductive content analysis methodology, employed in~\citet{values_encoded_in_ML}, and aim to follow best practices and recommendations from the qualitative content analysis literature \citep[see e.g.,][]{qualitative_content_analysis_4, qualitative_content_analysis_3, qualitative_content_analysis_2, qualitative_content_analysis_1}.

We start with a set of ten objectives described in \citet{zhong-etal-2020-nlp}, such as relation extraction, legal judgment prediction and question answering. 
We list all of these in Appendix \ref{initial_objectives}.
If a paper describes an objective that cannot be classified into one of these categories, we define a new category.
After every 20 annotations, we review all categories assigned so far and decide whether to merge existing categories (e.g., named entity recognition and semantic role labeling merge into information extraction, while similar case matching and semantic search merge into information retrieval) -- this represents the deductive step in our approach.
After annotating all 171 papers, we review all categories again and perform a final merge.
Finally, for each paper, we determine whether it is a legal, NLP, or interdisciplinary publication. 
This categorization is based on the publication venue and the author's affiliations.

We display our main findings in Figure \ref{fig:main_results}. 
We find that despite the inherently interdisciplinary nature of legal NLP, most work we reviewed is produced and consumed by the NLP community. 
Only 10\% of the papers citing \citet{zhong-etal-2020-nlp} are primarily interdisciplinary. 
Perhaps more strikingly, only a single law review article cites \citet{zhong-etal-2020-nlp}. 
In terms of paper objectives, we find that legal judgment prediction appears to be the most popular objective, with 20\% of reviewed papers focusing on this task. 

\section{Categorizing legal NLP applications}

Based on our review, we categorize popular legal NLP objectives in terms of their ability to impact the practice of law or broaden access to justice.
Subsequently, we identify potentially impactful areas of research that remain underexplored.

\subsection{Existing applications that  promise to aid legal practitioners}

We identify three primary streams of legal NLP research that promise to benefit legal practice and improve access to justice. 

\paragraph{Legal document generation and analysis.} NLP technology can help speed up drafting, e.g., via dedicated legal language models or general-purpose language models such as GPT-4. It can also help analyze legal documents, for example via information extraction or summarization~\citep{galgani2012combining, bommarito2021lexnlp}. Based on a survey of senior attorneys (see Appendix \ref{LLF_survey}), we note that document review and generation appear to be critical tasks from the perspective of many practitioners. 

\paragraph{Semantic search.} Legal arguments rely heavily on citations to statutes and precedential court opinions. 
Several scholars have thus focused on designing systems that aid attorneys in finding citations to prior court decisions that support their arguments~\cite{huang2021context, tang2021searching}. 

\paragraph{Accessibility of legal language.} The translation of legal jargon into more accessible forms has been identified as an important priority by legal scholars~\cite{benson1984end}. 
Here, style transfer methods, legal summarization, question answering and information extraction methods can all prove helpful to make legal language more accessible and to surface key concepts~\citep[see e.g.,][]{farzindar-lapalme-2004-legal, manor-li-2019-plain, khazaeli-etal-2021-free}. 
These applications can help judges quickly understand filings submitted by attorneys, aid lawyers in gaining an overview of documents, and can help individuals better understand contracts, wills and other documents that may be relevant to them.

\subsection{Applications that fail to aid legal practitioners}

Some legal NLP publications focus on tasks that are simply not part of legal practice or that use legal data in ways that do not fully account for how this data was generated. 
Other publications focus on tasks with significant ethical implications that make them ill-suited for real-world deployment.

Legal judgment prediction (LJP), the most common task identified in our review, suffers from both of these weaknesses.
First, LJP typically extracts facts from court opinions and then uses the facts to predict the associated judgment. 
This approach is problematic because the narrative presented by judges in their opinions is typically crafted with an outcome in mind, thereby precluding neutrality in the facts they present.
As such, LJP treats human-generated annotations as ground truths when in fact these annotations are based on confounding factors. 
Moreover, the automation of legal judgments is fraught with ethical challenges.
Biased judgments would have grave social implications, not only for litigants directly affected by inaccurate legal judgments but also society at large if automated judgments undermine trust in the judicial system.
LJP may have utility for low-stakes disputes that may not otherwise see a day in court, or it could be used to simulate a specific judge's or court's idiosyncrasies, which may be a helpful strategic tool for potential litigants. 
Furthermore, LJP might also be useful to surface existing biases in judicial decision making. 
However, LJP is typically framed as modeling the ``correct'' application of laws to facts. 
Due to its inherent risks, this application should be carefully considered and it is unlikely to materialize in the near future, if at all.

It is important to underscore that other common legal NLP tasks may not directly aid legal practitioners, but nevertheless provide valuable resources and insights. 
These include detecting bias in legal language \cite{bias_in_legal_language} and legal NLP benchmarks which help measure the progress of NLP methods~\cite{chalkidis-etal-2022-lexglue,guha2023legalbench}.

\subsection{Underexplored applications that promise to aid legal practitioners}

Understanding the nature of legal practice in more detail can help surface applications of NLP that would be useful to legal practitioners.
Of course, this is easier said than done as there are still limited opportunities for legal NLP researchers and legal practitioners to exchange ideas.
For this discussion, we draw partially on a survey conducted as part of Harvard Law School's 2023 \textit{Leadership in Law Firms} program (LLF).
This survey asked over 50 senior attorneys from 17 different countries to identify potentially impactful applications of NLP which would provide value in their firms (see Appendix \ref{LLF_survey} for an overview of responses).\footnote{We recognize that these responses are not representative of legal practice generally, but present them as a valuable example of how practitioners think about NLP and as a starting point for ideation.}

\paragraph{Persuasive legal reasoning.} Litigation is at least partially a rhetorical exercise in which attorneys seek to identify the most persuasive arguments while taking into account the presiding judge and, in some instances, the composition of the jury.
The nature of litigation offers ample opportunity for the study of language in general, and the study of discourse and pragmatics specifically.
Extraneous factors, like the presiding judge, have a significant impact on the persuasiveness of different arguments, and there already exists NLP research on context-aware argumentation~\citep[see e.g.][]{durmus2019role} that could be applied to law.

\paragraph{Practice-oriented legal research tools.} Legal research and case analysis was one of the key areas identified in the LLF Survey.
In common law jurisdictions, law develops organically through judicial opinions and higher courts may overturn or refine past court decisions.
Legal research platforms label whether a case is considered ``good law'', that is whether it remains a good basis for future arguments and thus current law.
Current citation prediction work has largely ignored this aspect, creating a risk that outdated or overturned opinions are recommended. 
NLP research techniques such as sentiment analysis could identify \textit{how} a citation is used by judges to determine whether it remains good law.

A related extension is the retrieval of individual legal passages. 
Judicial opinions are generally long documents and legal practitioners normally cite very specific passages.
As a result, legal research platforms often present specific passages as ``head notes'' or ``key cites'' to allow lawyers and judges to identify the most important portions of opinions.
Legal passage prediction (LPP) seeks to predict specific passages, rather than entire judicial opinions, which is more closely aligned with the needs of legal professionals~\cite{mahari2021autolaw}.
LPP may also be combined with extractive summarization~\citep[see e.g.][]{bauer2023legal}, to identify passages from an opinion that are most likely to represent useful citations.

\paragraph{Retrieval augmented generation over private legal data.} A more general opportunity for legal NLP is related to proprietary legal data.
Law firms amass large knowledge banks from past cases that contain sensitive and confidential data.
Practicing attorneys routinely build on their past work and experience. NLP tools could help them identify relevant records and, based on these retrieved records, generate new documents. 
Retrieval augmented generation~\citep[see e.g.][]{lewis2020retrieval, pmlr-v162-borgeaud22a, shi2023replug} is well suited to this task, however, it is critical that confidential records are not leaked to external parties or other public databases~\cite{arora2023reasoning}, and that generation is performed in an auditable fashion~\cite{mahari2023transparency}.

\section{Discussion}

The disconnect between AI research on applications and specific disciplines is not limited to law~\citep[see e.g.][]{acosta2022multimodal}.
Law, however, is unique among disciplines in that it is a field built on language.
Given the current state of legal practice, there is a need for innovation to make legal services more affordable and to address the access to justice crisis.
As such, law presents a valuable opportunity for the NLP community to conduct research on applications that could aid legal practitioners and that expand access to justice.

Impactful legal NLP research must be grounded in the needs of the legal community. 
The observed lack of cross-disciplinary citations in our review suggests that legal NLP researchers are largely disconnected from the legal community.
We encourage legal NLP researchers to identify tasks that are performed frequently by legal practitioners and that lend themselves to the application of NLP techniques.
To aid NLP researchers in identifying these tasks, we urge them to consider closer interdisciplinary collaborations with the legal community or at least to address legal issues identified in the legal literature.

\section{Conclusion}

By leveraging a literature review, we find that the legal NLP community is largely disconnected from legal academia. 
We emphasize that certain popular legal NLP tasks are only of limited utility to legal practitioners. We thus urge legal NLP researchers to focus on access to justice as a shared normative objective, to ground their work in the needs of the legal community, and to consider collaborating with lawyers to ensure that their research has applications in practice.
NLP has the potential to positively transform the practice of law and by extension society. However, this is impossible without cross-disciplinary understanding and collaboration.

\section*{Limitations}

\paragraph{Business applications.} 
Reviews of NLP literature provide insight into academic work, but they do not reveal business applications of NLP.
While we observe a disconnect between NLP research and law in academia, it is possible that there exists unpublished work that is more attuned to the needs of the legal community.
However, this work tends to focus on profitable applications, which are not always aligned with broadening access to justice.
The LLF survey provides a business-oriented perspective by surveying senior attorneys from an international group of law firms, however, more exhaustive work is needed to fully understand where NLP tools provide value to law firms and to what degree these offerings also address access to justice issues.

\paragraph{Scope.} We conduct a rapid review based on citations to a popular NLP paper. Our intuitions about the field lead us to believe that our findings extrapolate to the field as a whole. Contemporaneous work provides a broader overview and identifies similar trends as our review. 
For example, \citet{katz2023natural} find that classification is the most popular objective in the legal NLP community (LJP represents a classification task). 
While our work is similar in spirit to \citet{katz2023natural}, we take a less descriptive but more normative approach.

\section*{Ethics Statement}

If aligned with societal needs, legal NLP has tremendous potential to expand access to justice, to reduce biases in legal practice, and to create new efficiencies in legal practice.
At the same time, legal NLP deals with a sensitive aspect of society. 
Poorly designed NLP tools could embed biases, remove human oversight, or undermine trust in the legal system.
Our hope is that encouraging collaboration between NLP and legal researchers will also help identify and mitigate ethical challenges.

\paragraph{Broader Impact.} This publication is a perspective about the current state of legal NLP and its future directions, grounded in evidence about interdisciplinary disconnects. Of course, the trajectory of an academic field ought to be based on deliberative discussions involving many stakeholders. We present our recommendations and visions about the future of legal NLP, which are, at least to some extent, subjective. We invite others to expand on and to critique our views and hope to contribute to a broad and thoughtful discussion about the future of legal NLP.

\section*{Acknowledgements}
We are grateful to the Harvard Law School Center on the Legal Profession, Scott Westfahl, and David Wilkins for helping us gain insight into the needs of legal practitioners; we appreciate all participants in the Law Firms Leadership Program for taking time to share their perspectives with us. 
We thank Ron Dolin for his extremely helpful feedback on this paper.  

\bibliography{anthology,custom}

\begin{thebibliography}{57}
\expandafter\ifx\csname natexlab\endcsname\relax\def\natexlab#1{#1}\fi

\bibitem[{Acosta et~al.(2022)Acosta, Falcone, Rajpurkar, and
  Topol}]{acosta2022multimodal}
Juli{\'a}n~N Acosta, Guido~J Falcone, Pranav Rajpurkar, and Eric~J Topol. 2022.
\newblock Multimodal biomedical {AI}.
\newblock \emph{Nature Medicine}, 28(9):1773--1784.

\bibitem[{Arora et~al.(2023)Arora, Lewis, Fan, Kahn, and
  R{\'e}}]{arora2023reasoning}
Simran Arora, Patrick Lewis, Angela Fan, Jacob Kahn, and Christopher R{\'e}.
  2023.
\newblock Reasoning over public and private data in retrieval-based systems.
\newblock \emph{Transactions of the Association for Computational Linguistics},
  11:902--921.

\bibitem[{Bauer et~al.(2023)Bauer, Stammbach, Gu, and Ash}]{bauer2023legal}
Emmanuel Bauer, Dominik Stammbach, Nianlong Gu, and Elliott Ash. 2023.
\newblock Legal extractive summarization of {US} court opinions.
\newblock \emph{arXiv preprint arXiv:2305.08428}.

\bibitem[{Baxter(2022)}]{baxter2022dereliction}
Ralph Baxter. 2022.
\newblock Dereliction of duty: State-bar inaction in response to america's
  access-to-justice crisis.
\newblock \emph{Yale Law Journal Forum}, 132:228.

\bibitem[{Bengtsson(2016)}]{qualitative_content_analysis_1}
Mariette Bengtsson. 2016.
\newblock \href {https://doi.org/https://doi.org/10.1016/j.npls.2016.01.001}
  {How to plan and perform a qualitative study using content analysis}.
\newblock \emph{NursingPlus Open}, 2:8--14.

\bibitem[{Benson(1984)}]{benson1984end}
Robert~W Benson. 1984.
\newblock The end of legalese: The game is over.
\newblock \emph{New York University Review of Law \& Social Change}, 13:519.

\bibitem[{Birhane et~al.(2022)Birhane, Kalluri, Card, Agnew, Dotan, and
  Bao}]{values_encoded_in_ML}
Abeba Birhane, Pratyusha Kalluri, Dallas Card, William Agnew, Ravit Dotan, and
  Michelle Bao. 2022.
\newblock \href {https://doi.org/10.1145/3531146.3533083} {The values encoded
  in machine learning research}.
\newblock In \emph{Proceedings of the 2022 ACM Conference on Fairness,
  Accountability, and Transparency}, FAccT '22, page 173–184, New York, NY,
  USA. Association for Computing Machinery.

\bibitem[{Bommarito~II et~al.(2021)Bommarito~II, Katz, and
  Detterman}]{bommarito2021lexnlp}
Michael~J Bommarito~II, Daniel~Martin Katz, and Eric~M Detterman. 2021.
\newblock Lex{NLP}: Natural language processing and information extraction for
  legal and regulatory texts.
\newblock In \emph{Research Handbook on Big Data Law}, pages 216--227. Edward
  Elgar Publishing.

\bibitem[{Bommasani et~al.(2021)Bommasani, Hudson, Adeli, Altman, Arora, von
  Arx, Bernstein, Bohg, Bosselut, Brunskill
  et~al.}]{bommasani2021opportunities}
Rishi Bommasani, Drew~A Hudson, Ehsan Adeli, Russ Altman, Simran Arora, Sydney
  von Arx, Michael~S Bernstein, Jeannette Bohg, Antoine Bosselut, Emma
  Brunskill, et~al. 2021.
\newblock On the opportunities and risks of foundation models.
\newblock \emph{arXiv preprint arXiv:2108.07258}.

\bibitem[{Borgeaud et~al.(2022)Borgeaud, Mensch, Hoffmann, Cai, Rutherford,
  Millican, Van Den~Driessche, Lespiau, Damoc, Clark, De~Las~Casas, Guy,
  Menick, Ring, Hennigan, Huang, Maggiore, Jones, Cassirer, Brock, Paganini,
  Irving, Vinyals, Osindero, Simonyan, Rae, Elsen, and
  Sifre}]{pmlr-v162-borgeaud22a}
Sebastian Borgeaud, Arthur Mensch, Jordan Hoffmann, Trevor Cai, Eliza
  Rutherford, Katie Millican, George~Bm Van Den~Driessche, Jean-Baptiste
  Lespiau, Bogdan Damoc, Aidan Clark, Diego De~Las~Casas, Aurelia Guy, Jacob
  Menick, Roman Ring, Tom Hennigan, Saffron Huang, Loren Maggiore, Chris Jones,
  Albin Cassirer, Andy Brock, Michela Paganini, Geoffrey Irving, Oriol Vinyals,
  Simon Osindero, Karen Simonyan, Jack Rae, Erich Elsen, and Laurent Sifre.
  2022.
\newblock \href {https://proceedings.mlr.press/v162/borgeaud22a.html}
  {Improving language models by retrieving from trillions of tokens}.
\newblock In \emph{Proceedings of the 39th International Conference on Machine
  Learning}, volume 162 of \emph{Proceedings of Machine Learning Research},
  pages 2206--2240. PMLR.

\bibitem[{Bosio(2023)}]{bosio2023survey}
Erica Bosio. 2023.
\newblock A survey of judicial effectiveness: The last quarter century of
  empirical evidence.
\newblock Technical report, The World Bank.

\bibitem[{Cabral et~al.(2012)Cabral, Chavan, Clarke, and
  Greacen}]{cabral2012using}
James~E Cabral, Abhijeet Chavan, Thomas~M Clarke, and John Greacen. 2012.
\newblock Using technology to enhance access to justice.
\newblock \emph{Harvard Journal of Law \& Technology}, 26:241.

\bibitem[{Chalkidis et~al.(2022)Chalkidis, Jana, Hartung, Bommarito,
  Androutsopoulos, Katz, and Aletras}]{chalkidis-etal-2022-lexglue}
Ilias Chalkidis, Abhik Jana, Dirk Hartung, Michael Bommarito, Ion
  Androutsopoulos, Daniel Katz, and Nikolaos Aletras. 2022.
\newblock \href {https://doi.org/10.18653/v1/2022.acl-long.297} {{L}ex{GLUE}: A
  benchmark dataset for legal language understanding in {E}nglish}.
\newblock In \emph{Proceedings of the 60th Annual Meeting of the Association
  for Computational Linguistics (Volume 1: Long Papers)}, pages 4310--4330,
  Dublin, Ireland. Association for Computational Linguistics.

\bibitem[{Cunha et~al.(2018)Cunha, Gabbay, Ghirardi, Trubek, and
  Wilkins}]{cunha2018brazilian}
Luciana~Gross Cunha, Daniela~Monteiro Gabbay, Jos{\'e}~Garcez Ghirardi, David~M
  Trubek, and David~B Wilkins. 2018.
\newblock \emph{The Brazilian Legal Profession in the Age of Globalization}.
\newblock Cambridge University Press.

\bibitem[{Dale(2019)}]{dale_2019}
Robert Dale. 2019.
\newblock \href {https://doi.org/10.1017/S1351324918000475} {Law and word
  order: {NLP} in legal tech}.
\newblock \emph{Natural Language Engineering}, 25(1):211–217.

\bibitem[{de~Oliveira et~al.(2022)de~Oliveira, da~Silva~Gomes, Enes,
  Castelo~Branco, Pires, Bolzon, and Demo}]{deOliveira2022}
Leonardo~Ferreira de~Oliveira, Anderson da~Silva~Gomes, Yuri Enes,
  Tha{\'i}ssa~Velloso Castelo~Branco, Ra{\'i}ssa~Paiva Pires, Andrea Bolzon,
  and Gisela Demo. 2022.
\newblock \href {https://doi.org/10.1007/s43545-022-00482-w} {Path and future
  of artificial intelligence in the field of justice: a systematic literature
  review and a research agenda}.
\newblock \emph{SN Social Sciences}, 2(9):180.

\bibitem[{Durmus et~al.(2019)Durmus, Ladhak, and Cardie}]{durmus2019role}
Esin Durmus, Faisal Ladhak, and Claire Cardie. 2019.
\newblock The role of pragmatic and discourse context in determining argument
  impact.
\newblock In \emph{Proceedings of the 2019 Conference on Empirical Methods in
  Natural Language Processing and the 9th International Joint Conference on
  Natural Language Processing (EMNLP-IJCNLP)}.

\bibitem[{Eloundou et~al.(2023)Eloundou, Manning, Mishkin, and
  Rock}]{eloundou2023gpts}
Tyna Eloundou, Sam Manning, Pamela Mishkin, and Daniel Rock. 2023.
\newblock \href {http://arxiv.org/abs/2303.10130} {{GPT}s are {GPT}s: An early
  look at the labor market impact potential of large language models}.

\bibitem[{Fagan(2020)}]{fagan2020natural}
Frank Fagan. 2020.
\newblock Natural language processing for lawyers and judges.
\newblock \emph{Michigan Law Review}, 119:1399.

\bibitem[{Farzindar and Lapalme(2004)}]{farzindar-lapalme-2004-legal}
Atefeh Farzindar and Guy Lapalme. 2004.
\newblock \href {https://aclanthology.org/W04-1006} {Legal text summarization
  by exploration of the thematic structure and argumentative roles}.
\newblock In \emph{Text Summarization Branches Out}, pages 27--34, Barcelona,
  Spain. Association for Computational Linguistics.

\bibitem[{Galgani et~al.(2012)Galgani, Compton, and
  Hoffmann}]{galgani2012combining}
Filippo Galgani, Paul Compton, and Achim Hoffmann. 2012.
\newblock Combining different summarization techniques for legal text.
\newblock In \emph{Proceedings of the workshop on innovative hybrid approaches
  to the processing of textual data}, pages 115--123.

\bibitem[{Guha et~al.(2023)Guha, Nyarko, Ho, R{\'e}, Chilton, Narayana,
  Chohlas-Wood, Peters, Waldon, Rockmore et~al.}]{guha2023legalbench}
Neel Guha, Julian Nyarko, Daniel~E Ho, Christopher R{\'e}, Adam Chilton, Aditya
  Narayana, Alex Chohlas-Wood, Austin Peters, Brandon Waldon, Daniel~N
  Rockmore, et~al. 2023.
\newblock Legalbench: A collaboratively built benchmark for measuring legal
  reasoning in large language models.
\newblock \emph{arXiv preprint arXiv:2308.11462}.

\bibitem[{Hendrycks et~al.(2021)Hendrycks, Burns, Chen, and
  Ball}]{hendrycks2021cuad}
Dan Hendrycks, Collin Burns, Anya Chen, and Spencer Ball. 2021.
\newblock Cuad: An expert-annotated {NLP} dataset for legal contract review.
\newblock \emph{arXiv preprint arXiv:2103.06268}.

\bibitem[{Hsieh and Shannon(2005)}]{qualitative_content_analysis_2}
Hsiu-Fang Hsieh and Sarah~E. Shannon. 2005.
\newblock \href {https://doi.org/10.1177/1049732305276687} {Three approaches to
  qualitative content analysis}.
\newblock \emph{Qualitative Health Research}, 15(9):1277--1288.
\newblock PMID: 16204405.

\bibitem[{Huang et~al.(2021)Huang, Low, Teng, Zhang, Ho, Krass, and
  Grabmair}]{huang2021context}
Zihan Huang, Charles Low, Mengqiu Teng, Hongyi Zhang, Daniel~E Ho, Mark~S
  Krass, and Matthias Grabmair. 2021.
\newblock Context-aware legal citation recommendation using deep learning.
\newblock In \emph{Proceedings of the eighteenth international conference on
  artificial intelligence and law}, pages 79--88.

\bibitem[{Katz et~al.(2021)Katz, Dolin, and Bommarito}]{katz2021legal}
Daniel~Martin Katz, Ron Dolin, and Michael~J Bommarito. 2021.
\newblock \emph{Legal informatics}.
\newblock Cambridge University Press.

\bibitem[{Katz et~al.(2023)Katz, Hartung, Gerlach, Jana, and
  au2}]{katz2023natural}
Daniel~Martin Katz, Dirk Hartung, Lauritz Gerlach, Abhik Jana, and Michael J.
  Bommarito~II au2. 2023.
\newblock \href {http://arxiv.org/abs/2302.12039} {Natural language processing
  in the legal domain}.

\bibitem[{Khangura et~al.(2012)Khangura, Konnyu, Cushman, Grimshaw, and
  Moher}]{Khangura2012}
Sara Khangura, Kristin Konnyu, Rob Cushman, Jeremy Grimshaw, and David Moher.
  2012.
\newblock \href {https://doi.org/10.1186/2046-4053-1-10} {Evidence summaries:
  the evolution of a rapid review approach}.
\newblock \emph{Systematic Reviews}, 1(1):10.

\bibitem[{Khazaeli et~al.(2021)Khazaeli, Punuru, Morris, Sharma, Staub, Cole,
  Chiu-Webster, and Sakalley}]{khazaeli-etal-2021-free}
Soha Khazaeli, Janardhana Punuru, Chad Morris, Sanjay Sharma, Bert Staub,
  Michael Cole, Sunny Chiu-Webster, and Dhruv Sakalley. 2021.
\newblock \href {https://doi.org/10.18653/v1/2021.nllp-1.11} {A free format
  legal question answering system}.
\newblock In \emph{Proceedings of the Natural Legal Language Processing
  Workshop 2021}, pages 107--113, Punta Cana, Dominican Republic. Association
  for Computational Linguistics.

\bibitem[{Krippendorff(2018)}]{qualitative_content_analysis_3}
Klaus Krippendorff. 2018.
\newblock \emph{Content analysis: An introduction to its methodology}.
\newblock Sage publications.

\bibitem[{Laureate et~al.(2023)Laureate, Buntine, and Linger}]{doogan_review}
Caitlin Doogan~Poet Laureate, Wray Buntine, and Henry Linger. 2023.
\newblock A systematic review of the use of topic models for short text social
  media analysis.
\newblock \emph{Artificial Intelligence Review}, pages 1--33.

\bibitem[{Leivaditi et~al.(2020)Leivaditi, Rossi, and
  Kanoulas}]{leivaditi2020benchmark}
Spyretta Leivaditi, Julien Rossi, and Evangelos Kanoulas. 2020.
\newblock A benchmark for lease contract review.
\newblock \emph{arXiv preprint arXiv:2010.10386}.

\bibitem[{Lewis et~al.(2020)Lewis, Perez, Piktus, Petroni, Karpukhin, Goyal,
  K{\"u}ttler, Lewis, Yih, Rockt{\"a}schel et~al.}]{lewis2020retrieval}
Patrick Lewis, Ethan Perez, Aleksandra Piktus, Fabio Petroni, Vladimir
  Karpukhin, Naman Goyal, Heinrich K{\"u}ttler, Mike Lewis, Wen-tau Yih, Tim
  Rockt{\"a}schel, et~al. 2020.
\newblock Retrieval-augmented generation for knowledge-intensive {NLP} tasks.
\newblock \emph{Advances in Neural Information Processing Systems},
  33:9459--9474.

\bibitem[{Livermore and Rockmore(2019)}]{Livermore_Rockmore_2019}
Michael~A. Livermore and Daniel~N. Rockmore. 2019.
\newblock \emph{Law as Data: Computation, Text, and the Future of Legal
  Analysis}.
\newblock The Santa Fe Institute Press.

\bibitem[{Mahari(2021)}]{mahari2021autolaw}
Robert Mahari. 2021.
\newblock Autolaw: Augmented legal reasoning through legal precedent
  prediction.
\newblock \emph{arXiv preprint arXiv:2106.16034}.

\bibitem[{Mahari et~al.(2023)Mahari, South, and
  Pentland}]{mahari2023transparency}
Robert Mahari, Tobin South, and Alex Pentland. 2023.
\newblock Transparency by design for large language models.
\newblock \emph{Computational Legal Futures, Network Law Review}.

\bibitem[{Manor and Li(2019)}]{manor-li-2019-plain}
Laura Manor and Junyi~Jessy Li. 2019.
\newblock \href {https://doi.org/10.18653/v1/W19-2201} {Plain {E}nglish
  summarization of contracts}.
\newblock In \emph{Proceedings of the Natural Legal Language Processing
  Workshop 2019}, pages 1--11, Minneapolis, Minnesota. Association for
  Computational Linguistics.

\bibitem[{Martinez-Gil(2023)}]{martinez2023survey}
Jorge Martinez-Gil. 2023.
\newblock A survey on legal question--answering systems.
\newblock \emph{Computer Science Review}, 48:100552.

\bibitem[{Merriam and Grenier(2019)}]{qualitative_content_analysis_4}
Sharan~B. Merriam and Robin~S. Grenier. 2019.
\newblock \href {https://books.google.com/books?id=PL59DwAAQBAJ}
  {\emph{Qualitative Research in Practice: Examples for Discussion and
  Analysis}}.
\newblock Wiley.

\bibitem[{Nyarko and Sanga(2022)}]{nyarko2022statistical}
Julian Nyarko and Sarath Sanga. 2022.
\newblock A statistical test for legal interpretation: Theory and applications.
\newblock \emph{The Journal of Law, Economics, and Organization},
  38(2):539--569.

\bibitem[{OpenAI(2023)}]{openai2023gpt4}
OpenAI. 2023.
\newblock \href {http://arxiv.org/abs/2303.08774} {{GPT}-4 technical report}.

\bibitem[{Ostendorff et~al.(2021)Ostendorff, Ash, Ruas, Gipp, Moreno-Schneider,
  and Rehm}]{ostendorff2021evaluating}
Malte Ostendorff, Elliott Ash, Terry Ruas, Bela Gipp, Julian Moreno-Schneider,
  and Georg Rehm. 2021.
\newblock Evaluating document representations for content-based legal
  literature recommendations.
\newblock In \emph{Proceedings of the Eighteenth International Conference on
  Artificial Intelligence and Law}, pages 109--118.

\bibitem[{Pace et~al.(2023)Pace, Brink, Lee, and Hanlon}]{Pace2023}
Nicholas~M. Pace, Malia~N. Brink, Cynthia~G. Lee, and Stephen~F. Hanlon. 2023.
\newblock National public defense workload study.
\newblock Technical report, RAND Corporation.

\bibitem[{Rhode(2013)}]{rhode2013access}
Deborah~L Rhode. 2013.
\newblock Access to justice: A roadmap for reform.
\newblock \emph{Fordham Urban Law Journal}, 41:1227.

\bibitem[{Rice et~al.(2019)Rice, Rhodes, and Nteta}]{bias_in_legal_language}
Douglas Rice, Jesse~H. Rhodes, and Tatishe Nteta. 2019.
\newblock \href {https://doi.org/10.1177/2053168019848930} {Racial bias in
  legal language}.
\newblock \emph{Research \& Politics}, 6(2):2053168019848930.

\bibitem[{Ricketts et~al.(2023)Ricketts, Barry, Guo, and
  Pelham}]{slr_application}
Jon Ricketts, David Barry, Weisi Guo, and Jonathan Pelham. 2023.
\newblock \href {https://doi.org/10.3390/safety9020022} {A scoping literature
  review of natural language processing application to safety occurrence
  reports}.
\newblock \emph{Safety}, 9(2):22.

\bibitem[{Savelka et~al.(2019)Savelka, Xu, and Ashley}]{savelka2019improving}
Jaromir Savelka, Huihui Xu, and Kevin~D Ashley. 2019.
\newblock Improving sentence retrieval from case law for statutory
  interpretation.
\newblock In \emph{Proceedings of the seventeenth international conference on
  artificial intelligence and law}, pages 113--122.

\bibitem[{Shi et~al.(2023)Shi, Min, Yasunaga, Seo, James, Lewis, Zettlemoyer,
  and Yih}]{shi2023replug}
Weijia Shi, Sewon Min, Michihiro Yasunaga, Minjoon Seo, Rich James, Mike Lewis,
  Luke Zettlemoyer, and Wen-tau Yih. 2023.
\newblock Replug: Retrieval-augmented black-box language models.
\newblock \emph{arXiv preprint arXiv:2301.12652}.

\bibitem[{Slosar(2022)}]{LegalServicesCorporation2022}
Mary~C. Slosar. 2022.
\newblock The justice gap: The unmet civil legal needs of low-income americans.
\newblock Technical report, Legal Services Corporation.

\bibitem[{Tang and Clematide(2021)}]{tang2021searching}
Li~Tang and Simon Clematide. 2021.
\newblock \href {https://doi.org/10.18653/v1/2021.nllp-1.12} {Searching for
  legal documents at paragraph level: Automating label generation and use of an
  extended attention mask for boosting neural models of semantic similarity}.
\newblock In \emph{Proceedings of the Natural Legal Language Processing
  Workshop 2021}, pages 114--122, Punta Cana, Dominican Republic. Association
  for Computational Linguistics.

\bibitem[{Tsarapatsanis and Aletras(2021)}]{tsarapatsanis-aletras-2021-ethical}
Dimitrios Tsarapatsanis and Nikolaos Aletras. 2021.
\newblock \href {https://doi.org/10.18653/v1/2021.findings-acl.314} {On the
  ethical limits of natural language processing on legal text}.
\newblock In \emph{Findings of the Association for Computational Linguistics:
  ACL-IJCNLP 2021}, pages 3590--3599, Online. Association for Computational
  Linguistics.

\bibitem[{Vold and Conrad(2021)}]{vold2021using}
Andrew Vold and Jack~G Conrad. 2021.
\newblock Using transformers to improve answer retrieval for legal questions.
\newblock In \emph{Proceedings of the Eighteenth International Conference on
  Artificial Intelligence and Law}, pages 245--249.

\bibitem[{Wilkins et~al.(2017)Wilkins, Khanna, and Trubek}]{wilkins2017indian}
David~B Wilkins, Vikramaditya~S Khanna, and David~M Trubek. 2017.
\newblock \emph{The Indian Legal Profession in the Age of Globalization}.
\newblock Cambridge University Press.

\bibitem[{{World Justice Project}(2019)}]{WorldJusticeProject2019}
{World Justice Project}. 2019.
\newblock \href
  {https://worldjusticeproject.org/sites/default/files/documents/WJP-A2J-2019.pdf}
  {Global insights on access to justice: Findings from the world justice
  project general population poll in 101 countries}.

\bibitem[{Yang et~al.(2022)Yang, MacAvaney, Lewis, and
  Frieder}]{yang2022goldilocks}
Eugene Yang, Sean MacAvaney, David~D Lewis, and Ophir Frieder. 2022.
\newblock Goldilocks: Just-right tuning of {BERT} for technology-assisted
  review.
\newblock In \emph{European Conference on Information Retrieval}, pages
  502--517. Springer.

\bibitem[{Zhong et~al.(2020)Zhong, Xiao, Tu, Zhang, Liu, and
  Sun}]{zhong-etal-2020-nlp}
Haoxi Zhong, Chaojun Xiao, Cunchao Tu, Tianyang Zhang, Zhiyuan Liu, and Maosong
  Sun. 2020.
\newblock \href {https://doi.org/10.18653/v1/2020.acl-main.466} {How does {NLP}
  benefit legal system: A summary of legal artificial intelligence}.
\newblock In \emph{Proceedings of the 58th Annual Meeting of the Association
  for Computational Linguistics}, pages 5218--5230, Online. Association for
  Computational Linguistics.

\bibitem[{Zou and Kanoulas(2020)}]{zou2020towards}
Jie Zou and Evangelos Kanoulas. 2020.
\newblock Towards question-based high-recall information retrieval: Locating
  the last few relevant documents for technology-assisted reviews.
\newblock \emph{ACM Transactions on Information Systems (TOIS)}, 38(3):1--35.

\end{thebibliography}
\bibliographystyle{acl_natbib}

\appendix

\section{Initial Objectives}
\label{initial_objectives}

\begin{table}[!h]
    \centering
    \begin{tabular}{c}
Relation Extraction\\
Event Timeline\\
Element Detection\\
Legal Judgment Prediction\\
Question Answering\\
Similar Case Matching\\
Summarization\\
Embeddings\\
Knowledge Graphs\\
Language Models    \end{tabular}
    \caption{Set of initial objectives for the literature review.}
    \label{tab:initial_objectives}
\end{table}

\section{Harvard Leadership in Law Firms: Survey Responses}
\label{LLF_survey}
As part of Harvard Law School's \textit{Leadership in Law Firms} program, the lead author of this paper conducted a survey of over 50 senior attorneys from 17 different countries on applications of NLP in legal practice. The survey asked: \textit{What is one legal-related task (e.g., document review, responding to a motion), system (e.g., time entry or giving feedback), type of legal matter (deals, regulatory review) that you would LOVE for generative AI to make easier/more efficient?}

While by no means representative of the legal industry as a whole, the survey responses provide valuable insight into the priorities of practicing attorneys.
As such, they serve as a starting point for some new avenues of legal NLP research and an example of how the research community can solicit insights from practitioners.

At a high level, the following application categories emerged from the survey:
\begin{enumerate}
     [(1),left=0mm,topsep=0.1mm,noitemsep]
    \item Legal and business document review and summarization (42/59)
    \item Time entry and billing, case intake, and reviewing invoices (14/59)
    \item Case law research and regulatory review (11/59)
    \item Legal document generation, creating multi-document summaries (3/59)
    \item Simulating or predicting legal outcomes (2/59)
    \item Project and process management (2/59)
    \item Knowledge management (1/59)
\end{enumerate}

\noindent\textbf{Raw Responses}\\
\footnotesize
1. Document review and summaries\\
2. Case review intake\\
3. Initial research from multiple sources to create first draft memo. For financial services regulatory\\
4. Document review, precedents, process improvement\\
5. Time entry\\
6. Case analysis and statistics; usage in the discovery process\\
7. Billing Process\\
8. Time entry\\
9. Summarise an extensive document or prepare a well substantiated research\\
10. Financial/tax modelling, document review\\
11. Documentation review, regulatory review\\
12. Analysis of high volume procedural evidence\\
13. Document review\\
14. Time entry\\
15. Legal research (e.g., finding relevant case law)\\
16. Time recording\\
17. Predictive tool for client outcomes\\
18. Document review\\
19. Time entry\\
20. Documents review\\
21. Document and regulatory review\\
22. Document review in large, complex litigation\\
23. First draft of letters to media and social media\\
24. Review of trading data in securities enforcement matters\\
25. Time entry and billing\\
26. Legal research and regulatory option \\
27. Use of AI in the analysis of the firm's own data sets in order to make use of expertise available in the firm as quickly and effectively as possible, such as previously prepared expert opinions on a topic. Furthermore, it should be possible to search external databases as effectively as possible\\
28. Document review\\
29. Document review\\
30. Dealing with AML and KYC obligations\\
31. Document review\\
32. Time recording\\
33. Admin related tasks around time entry, review of accounts\\
34. One reliable source with brief and regular updates on case-law, legislation and important developments including access to in-depth information for the individual sources of law\\
35. Brief writing, time entry, legal research\\
36. Document Review and Legal Search - time entry - regulatory review\\
37. Lease summaries\\
38. Document review\\
39. Time entry\\
40. Giving feedback\\
41. Automating tasks/workflows in the sense of having a spread sheet/document assistant\\
42. Document review in diligence processes\\
43. Time keeping\\
44. Document Review\\
45. Document review (both consulting and litigation)\\
46. Document review\\
47. Document review\\
48. A system providing full and reliable overviews on legal topics by analyzing all relevant sources including legislation, legislation processes, case law and literature. This would help to often spend long time on getting certainty about being up to date\\
49. Document review\\
50. Comparative summary of publications and judgments\\
51. Horizon scanning for regulatory change\\
52. Document review\\
53. Analysis and comparison of different information sources, Intranet, Internet, databases\\
54. Document review in both counseling and litigation\\
55. Document review, regulatory review\\
56. Document Review\\
57. Legal-related task: document review and research; System: project management; Legal Matter: Due Diligence/deals\\
58. Administrative things like time entry or reviewing invoices\\
59. Summarise big volume of data

\end{document}